# Network Topology and Time Criticality Effects in the Modularised Fleet Mix Problem

**James Whitacre[1]; Axel Bender[2]; Stephen Baker[2]; Qi Fan[1]; Ruhul Sarker[1]; Hussein A. Abbass[1]**

[1] Defence and Security Applications Research Centre, UNSW@ADFA, Canberra, Australia
[2] Defence Science and Technology Organisation, Edinburgh, Adelaide, Australia
Email: j.whitacre@adfa.edu.au, axel.bender@dsto.defence.gov.au, steve.baker@dsto.defence.gov.au, q.fan@adfa.edu.au, r.sarker@adfa.edu.au, h.abbass@adfa.edu.au,

**Abstract.** In this paper, we explore the interplay between network topology and time criticality in a military logistics system. A general goal of this work (and previous work) is to evaluate land transportation requirements or, more specifically, how to design appropriate fleets of military general service vehicles that are tasked with the supply and re-supply of military units dispersed in an area of operation. The particular focus of this paper is to gain a better understanding of how the logistics environment changes when current Army vehicles with fixed transport characteristics are replaced by a new generation of modularised vehicles that can be configured task-specifically. The experimental work is conducted within a well developed strategic planning simulation environment which includes a scenario generation engine for automatically sampling supply and re-supply missions and a multi-objective meta-heuristic search algorithm (i.e. Evolutionary Algorithm) for solving the particular scheduling and routing problems. The results presented in this paper allow for a better understanding of how (and under what conditions) a modularised vehicle fleet can provide advantages over the currently implemented system.

## 1. INTRODUCTION

Designing the best deployable Field Vehicles, Modules, and Trailers (FVM&T) fleet(s) involves the non-trivial problem of identifying a suitable fleet mix that will meet future operational and strategic requirements. The future FVM&T fleet examined in this paper makes use of a vehicle 'modularisation' concept which sees a truck-trailer combination configurable to perform specific tasks or functions through the addition of appropriate 'modules'. This results in the separation of payload task functionality from the movement and manoeuvre functions of the base vehicle or trailer. Vehicles of a cargo functional type, for example, are created by adding a container, bulk liquid tank or dump module to a motive unit, i.e. the vehicle chassis. With interfaces common across a range of vehicles and trailers, the system components are designed such that modules can easily be swapped to meet contemporary mission requirements. The concept of modularisation adds a novel dimension to the problem of FVM&T fleet design and the intention in this paper is to investigate some of the unique challenges and opportunities afforded by a modularised fleet.

As a result of the complexity inherent to the FVM&T system in both the spectrum of its functions and the breadth of contributions it makes during military operations, it is insufficient to optimise the future FVM&T fleet against a single performance measure. Instead a number of objectives (which are often in conflict with each other) are necessary in order to adequately define the optimization environment. When resolving the FVM&T fleet design in terms of these multiple objectives it is important to explore and understand the trade-offs that are being made when implementing different FVM&T options.

The Modularised Fleet Mix Problem (MFMP) addressed in this work was proposed and studied in previous work [1], [2], [3] and can be defined as:

*A deployed military force has a range of mobility tasks to be undertaken utilising a heterogeneous modularised vehicle fleet incorporating truck and/or trailer operations. The deployed military force along with its vehicle fleet is distributed among many locations in an area of operation. Each mobility task is characterised as requiring a number and range of modules to be moved between locations, to meet a priority for movement, within time window constraints. Each truck and trailer type has characteristics in terms of its ability to carry a particular range and number of modules and its ability to move across particular terrain classifications at particular speeds. The problem is to select trucks, trailers and modules assets so as to provide the best fleet outcomes.*

A particular MFMP scenario therefore must describe the range of mobility tasks to be performed in a military setting. Here the military setting is based on the routing topology that defines feasible travel routes, supply sizes, supply types, supply replenishment cycles, and flexibility in the delivery time of supplies.

To develop solutions to the MFMP, we use a multi-objective multi-agent solver that has been developed specifically for the MFMP problem and is used to find a suitable fleet mix. This solver tackles a number of problems including scheduling, routing and binning as well as the coordination of fleet management, route planning and load/unload functionality. The solution from this process is a set of feasible fleet mixes (see Section 2 below).

**MFMP attributes under study**: This paper investigates FVM&T dependence on a selected set of problem parameters, namely network topology and parameters controlling time-criticality of FVM&T tasks. These parameters were specifically selected because they are known to play an important role in military operations and are furthermore expected to have a significant influence on the operational impact of vehicle modularisation.

The next section briefly reviews the multi-objective solver that is used to generate fleet mix solutions which is followed by a description of MFMP simulations in Section 3. The experimental results are presented and described in Section 4 with a summary of conclusions in Section 5.

## 2. MFMP FORMULATION

The MFMP is formulated as a constrained multi-objective problem. The MFMP problem constraints include:

- spatial, temporal and load constraints associated with mobility tasks within a problem instance (e.g. task origin/destination, time window for task commencement and completion, the nature and quantity of the loads, the compatibility / incompatibility of load types);

- constraints related to the operation of each vehicle, module and trailer type within the heterogeneous fleet (e.g. allowable weight and size of loads; permissible combinations of vehicle, trailer and module);

- conditions relating to employed personnel (e.g. allowable working hours, size of crew, driver competency requirements, rest requirements);

- conditions dictated by the environmental and tactical setting of the scenario (e.g. trafficable routes and restrictions to accessibility of nodes, convoy requirements).

Some of the constraints are specific to the military context (e.g. enemy threat, mission command) and can change over time as the military operation unfolds. Some of these dynamics are captured within individual MFMP problem instances by defining time dependent constraints which change over the duration of a scenario.

Formally the MFMP objectives are described as conditions on the $n$-tuple ($V^{k,t}, T^k, M^r$). Here $V^{k,t}$ and $T^k$ denote the number of vehicles and trailers of type $k$, respectively. The index $k$ is a vector index that represents both the vehicle/trailer type and the combination of modules that the vehicle/trailer motive unit can carry. The index $t$ encodes the vehicle-trailer combinations that are permissible including the potential of one vehicle pulling more than one trailer (if permissible). In the abstract scenario space explored in this study we ignore the problem domain of trailers; thus we set $T^k$ to zero and drop the index $t$ in the formulae to follow.

The term $M^r$ describes the number of modules, and the vector index $r$ denotes both the module type and the set of materials/equipment/people that are compatible with each module. Let $C^k$ be the cost of a vehicle of type $k$, $L^k$ the length of the vehicle, and $C^r$ the cost of a module of type $r$. Here the cost term refers to the total expected lifetime cost of a vehicle including purchasing cost, maintenance, and operational costs. The problem then is to identify a mixture of vehicles, modules and trailers to fulfill all tasks such that:

- The cost is minimum,

$$\min \left\{ \sum_k C^k V^k + \sum_r C^r M^r \right\}. \quad (1)$$

- The FVM&T vehicle mix is diverse;

$$\min \frac{1}{|k|} \sum_k \left( V^k - \overline{V} \right)^2. \quad (2)$$

- The space a vehicle occupies in a strategic sealift vessel is minimum,

$$\min \sum_k L^k V^k. \quad (3)$$

For the objective defined in (2), $|k|$ is the cardinality of the set $k$ while $\overline{V}$ is the average number of vehicles of all types. This objective maximises fleet diversity so that the fleet composition is not dominated by a single vehicle type. It is an operational heuristic based on extensive military experience which has indicated a positive correlation between operational flexibility and the diversity of asset types within a theatre of operation.

The objective defined in (3) is a surrogate for the cost of strategic lift (i.e. the cost for deploying FVM&T units to an area of operation). This objective is measured in *lane meters* which approximates the length of these pieces of equipment if they were arranged in a straight line. The significance of this objective stems from the limited resources that a military operation has for shipping large pieces of equipment to overseas destinations. Here we make the simplifying assumption that, during the strategic lift, all modules are attached to motive units and thus do not add to the strategic lift 'bill'. We also assume that shipment size constraints are violated before weight constraints which is almost always valid for the vehicle types considered here.

Typically, these three objectives are in conflict, i.e. solutions that optimise one of the objectives may be sub optimal when assessed against the other objectives. To resolve this conflict we apply a relaxed version of the Pareto dominance concept and generate a set of FVM&T options where each option is equivalent to or better (non-dominated) than any other option in the set in at least one of these three objectives. For more information on the concept of Pareto dominance and multi-objective optimisation, we refer the reader to [4].

**Solver Description**: The multi-objective problem is wrapped into a multi-agent solver environment which is described in detail in our previous work [2], [3]. Key agents in this environment are the Task Manager which heuristically selects mobility tasks for scheduling, the Fleet Manager which decides on the vehicle-module-trailer combination that is selected for a given task, the Schedule/Routing Manager which considers many of the problem conditions described previously to schedule the tasks and heuristically selects a permissible route. The solver applies an evolutionary approach, which, in each step, generates a set of FVM&T options. It then selects and recombines these options, repairs those FVM&T systems generated through recombinant operations and applies the Pareto dominance concept to make a decision on whether to keep or discard the newly created solution. It is worth noting that every generated solution is designed to meet all stated problem constraints and are feasible with respect to the successful completion of all tasks within a given problem instance.

After sufficiently many iterations of this process a comprehensive set of non-dominated solutions is evolved which form the input to our study on network topology and time-criticality dependence on fixed and modular fleets. A detailed description of the solver as a multi-agent system can be found in [2], [3].

Although the system being studied is solved within the multi-objective environment just described, to easily visualize results and report on those scenario features impacting modularised fleet performance requires us to obtain a single representative solution for each problem instance. This is addressed by presenting solution results that were best able to minimise the cost objective. It is important to emphasise however that the solutions generated accounted for each of the objectives stated in the problem definition.

## 3. PROBLEM INSTANCE DEFINITION AND MFMP SIMULATION DYNAMICS

This section describes the features of a single problem instance. Other examples of defence logistics models in the literature can be found in [5] and [6].

### 3.1 Task Characteristics

**Task definitions**: As previously mentioned, an MFMP involves the execution of a set of mobility tasks. In this study, two unique task types are considered, each with two possible sizes: 'medium' or 'heavy'. Tasks of type-1 can only be carried by a type-1 fixed vehicle or a type-1 module, tasks of type-2 only by a type-2 vehicle or module. Medium-sized tasks fit onto a medium truck or a heavy truck; heavy tasks can only be fulfilled by a heavy truck. Also, a heavy truck can carry two medium-sized loads.

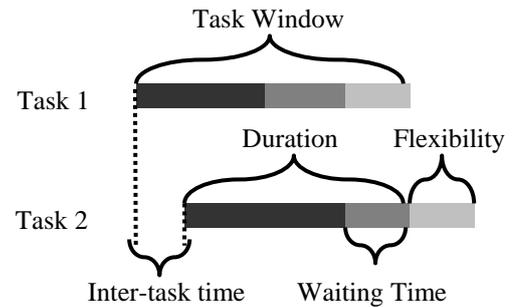

Figure 1 Definition of task duration, task window, waiting time, flexibility, and inter-task time

**Temporal parameters**: To define the sequence of tasks in a scenario, we introduce several temporal parameters as depicted in Figure 1. The inter-task time is a control parameter for task concurrency: the shorter the inter-task time the more likely it is that tasks occur concurrently. In particular, the earliest start time for a new task is uniformly sampled to be between zero and inter-task time minutes after the earliest start of the previous task.

The flexibility parameter is introduced to study the impact of time-criticality: the more flexible a task the less time-critical it is. Finally, the waiting time parameter is used to represent the additional loading/unloading time required when using fixed fleet vehicles compared to modularised fleet vehicles. This additional time is due to the fact that modules can be quickly loaded/unloaded onto a vehicle without needing to account for the movement of goods within the module.

### 3.2 Network Characteristics

Three network topologies are considered in these experiments. The first is a ten node ring structure (ring) which is used as a starting point for generating all other networks. The other networks, small-world 1 (SW1) and small-world 2 (SW2), are created by executing the topology morphing algorithm (below) one and two times, respectively. To promote higher centrality in SW2, node B from the first iteration of the topology morphing algorithm is selected to be either node A or node B in the second iteration. Examples of the resulting networks are provided in Figure 2. Comprehensive reviews on network morphing are not available in the literature although some information is provided in section 7.3 in [7].

**Topology morphing algorithm**

- select two nodes, A and B, at random
- copy all links from A and transfer to B
- remove links from A
- add a link between A and B

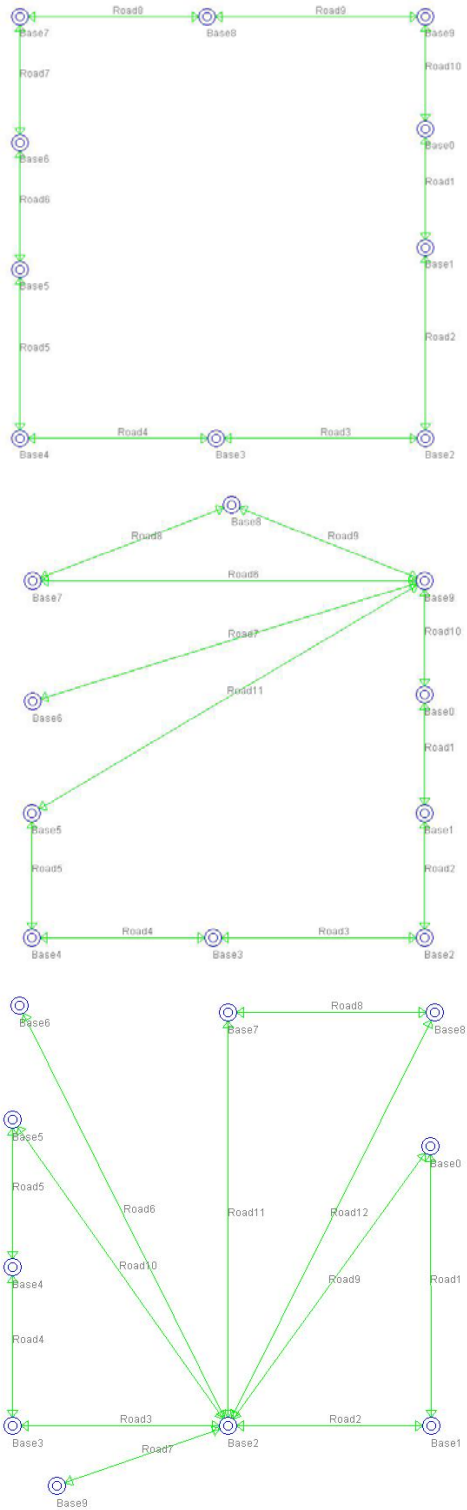

**Figure 2: Example networks for the Ring network (top), SW1 (middle), and SW2 (bottom)**

The topology morphing algorithm acts to create shortcuts in the available routing options however it can also lead to traffic bottlenecks in real transportation systems. The impact of the morphing algorithm on two topological properties of the networks, the characteristic path length L and the highest traffic node B, is shown in Table 1. The characteristic path length L measures the average shortest distance between nodes and indicates the average distance that vehicles will need to travel. The highest traffic node metric B measures the largest number of shortest paths within the network which travel through a single node. Small values for L roughly indicate a small-world effect in the network while large values for B indicate a highly centralised network. For a review of network topological properties, we refer the reader to [7], [8].

**Table 1: Network Properties**

| Network | L | B |
|---|---|---|
| Ring | 2.78 | 20.0 |
| SW1 | 2.36 | 51.9 |
| SW2 | 1.80 | 69.8 |

Within each network, tasks are randomly assigned origin and destination nodes. The vehicle or motive unit completing a task returns to its origin node after the task is completed. Modules are assumed to be returned to the origin node by the next available delivery vehicle however this is assumed to incur a negligible cost and is neglected in the simulation.

### 3.3 Vehicle Characteristics

In the fixed fleet mix, there are four vehicle types: two medium trucks for each of the two different task types and two heavy trucks. In the modularised fleet, there are two vehicle types (heavy/medium) and two module types (one for each task type). For the purpose of our analysis we make the assumption that modularisation does not incur a significant additional procurement cost; i.e. the cost of a medium- or heavy-sized vehicle-module combination in the modularised fleet is set equal to the cost of respectively the medium- or heavy-sized vehicle in the fixed fleet. While in reality there are cost differentials (e.g. because of cost differences between forklifts for the loading/unloading of vehicles in the fixed fleet and load handling systems that are integrated in the modularised vehicles), in most cases these are negligible compared with vehicle procurement costs.

### 4. RESULTS

#### 4.1 Experiment 1

The first set of experiments focuses on the impact of network topology on the cost of the FVM&T. In particular we are interested in understanding under what conditions a centralised, small-world network might provide a more cost-effective solution and when this type of network might be detrimental.

The obvious benefit from a small-world effect is that the average shortest path length and hence the average traveling distance for vehicles is much smaller. This implies that less vehicles might be needed to accomplish the same number of tasks. On the other hand, the presence of high traffic hubs in a network can cause congestion and delays in travel time. In a military context, owing to security concerns high traffic routes are often avoided. In particular, in military operations (including some supply missions) it is important to select routes in a fashion that is unpredictable to any adversary.

To investigate this tradeoff, we introduce a penalty to the cost function for vehicles each time they cross over a high traffic node. In particular, for each particular route considered by the Routing Agent, we count the number of vehicles that have crossed each node along the route in the last $y$ minutes and select the node with the highest traffic. A penalty is then added to that route which is equal to the amount of traffic on the highest traffic node times 0.1% of the vehicle purchase cost.

**MFMP Conditions**:

For all experiments the time to travel between two directly connected nodes is set to 30 minutes, the inter-task time is set to 5 minutes, and the total simulation time span is defined as 12 hours (for an average of roughly 3000 tasks generated per problem instance/simulation).

A task's duration is defined as the time required to travel over the longest path between origin and destination nodes. This ensures that each route is a feasible option which enables us to investigate the impact of creating shortcuts in the routing network (e.g. by using the topology morphing algorithm). The flexibility in the task start time is then defined as the time difference between any selected route and the longest route, meaning that for the longest route flexibility is set to 0.

Due to stochastic sampling of problem dynamics (e.g. inter-task time, origin and destination of tasks, task sizes, etc), 20 experimental replicates are conducted for each set of experimental conditions, with results shown as the average from these experiments. The experimental conditions which are varied include the network topology, the use of modular or fixed fleet, and the setting of $y$ which controls the importance of avoiding high traffic routes.

**Solver Conditions**: The multi-objective solver uses a population size of 20 and evolves for 100 generations to produce a non-dominated set of solutions for each problem instance. Given the other experimental conditions stated, this means for example that the results presented in Figure 3 are taken from 1.2 million simulations. As previously mentioned, the solutions represented in these results are those solutions with the lowest cost function as defined in equation (1).

The results for different values of $y$ are shown in Figure 3. Here we can see that when the traffic penalty $y$ is sufficiently small, the solution quality is better for networks with lower path lengths (i.e. cost of Ring > SW1 > SW2). However, as we place greater emphasis on avoiding high traffic routes, we find the preference order reverses so that a less centralised network is preferred (i.e. cost of Ring < SW1 < SW2). This result is observed regardless of whether the modular fleet or the fixed fleet is used. We can also see from these results that the modular fleet provides for a more cost effective solution under all experimental conditions which is likely a consequence of the more rapid unloading time associated with the modular fleet.

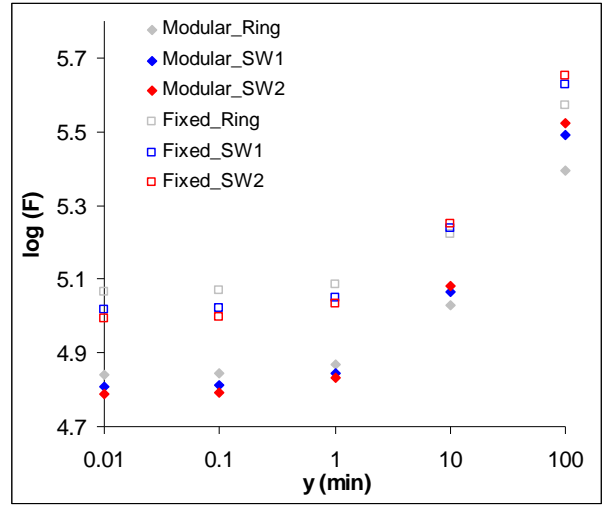

**Figure 3: Cost function (F) vs. penalty from high traffic routes.**

### 4.2  Experiment 2

In the second set of experiments, we investigate the influence of the flexibility parameter. Here we remove the penalty term related to travelling over high traffic routes (i.e. $y=0$) and flexibility is now varied as a proportion of each task's duration. All other experimental conditions are kept the same as those in the previous experiments.

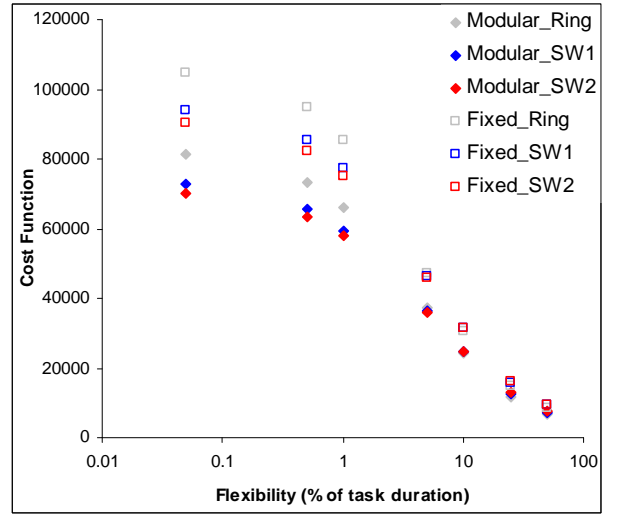

**Figure 4: Cost function vs. task scheduling time flexibility**

The results shown in Figure 4 display a number of interesting trends that highlight the interplay between topology, modular vs. fixed fleets, and the time criticality of tasks. First, one will notice that for all experimental conditions we observe a decreasing cost function for increasing values of flexibility. This is not surprising and simply indicates that the scheduling agent is able to utilise vehicle assets more efficiently as task execution time constraints are relaxed. However, this is not a simple linear relationship and instead we find only small changes to the cost function when flexibility is extended to very small or very large values. For large flexibility values this trend is less obvious in Figure 4 however we have run some tests with flexibility values greater than 100 and found the cost function flattens out to a stable value with fixed fleets a little less than double the cost of the modular fleets. This sustained additional cost reflects the fact that the fixed fleet requires two different vehicle types for each of the two task types while the modular fleet vehicles can be used for either task.

When measuring the modular fleet cost as a percentage of the fixed fleet under identical conditions, we were surprised to find that this value was not sensitive to flexibility for the range of values shown in the figure and remained at a constant value of 78% ($\pm 0.7\%$). Since the modular fleet vehicle can be utilised on a range of task types, we anticipated that this fleet would more efficiently utilise its assets when sufficient flexibility in task execution was provided. Future work will investigate why this behaviour was not observed.

Another interesting trend observed in Figure 4 is that the cost function's sensitivity to network topology is quickly lost as flexibility is increased in the system. This indicates that the advantages obtained by having shorter routes becomes less important as scheduling flexibility is increased.

In summary, the results presented in this paper indicate that:

- Highly centralised network topologies can provide for shorter routes however the advantages of this can be offset by an increasing risk from route predictability and more generally can result in traffic flow bottlenecks. Although the routing topology is not something that can normally be controlled within a theatre of operation, knowledge of its impact on operational effectiveness can help decision makers to anticipate and plan for potential challenges.

- The impact of network topology is dependent on the time criticality of tasks such that for highly flexible schedules, the small-world effect has an insignificant impact on solution quality. This could be particularly relevant in a military setting since task flexibility can be seen as a surrogate for enemy threat level: the higher the threat in a given scenario, the less likely it is that tasks have associated schedule flexibility.

- Flexibility has a significant impact on fleet size over a fairly broad range of values. Only at very small or very large values does task flexibility display a negligible impact on the cost of the FVM&T.

- A modularised fleet can provide cost savings over the fixed fleet which is largely derived from a faster unloading time. However, the benefits from the multi-functionality of modular vehicle assets were not observed under the experimental conditions tested.

## 5. CONCLUSION

In this paper, we have presented a series of experiments that were designed to study the effect of network topology and time criticality on the size and mix of general-service vehicles that are deployed in a military setting. They form part of a full analysis for the advantages and disadvantages of new concepts such as modularisation in military logistic settings. We have compared modularised and fixed military FVM&T systems and found that modularisation can markedly reduce the size of a deployed fleet, although the relative cost savings were not found to have a significant dependence on task flexibility over the range of experimental conditions tested. This latter behaviour is unexpected and will be subject of future research.